# A FRAMEWORK FOR CONTROL STRATEGIES IN UNCERTAIN INFERENCE NETWORKS

Moshe Ben-Bassat and Oded Maler

Faculty of Management         and        University of CAlifornia
Tel Aviv University                      Computer Science Department
tel Aviv 69978                           Los Angeles, California 90024
ISRAEL                                   U. S. A.

## Abstract

Control Strategies for hierachical tree-like probabilistic inference networks are formulated and investigated. Strategies that utilize staged look-ahead and temporary focus on subgoals are formalized and refined using the Depth Vector concept that serves as a tool for defining the 'virtual tree' regarded by the control strategy. The concept is illustrated by four types of control strategies for three-level trees that are characterized according to their Depth Vector, and according to the way they consider intermediate nodes and the role that they let these nodes play.

INFERENT1 is a computerized inference system written in Prolog, which provides tools for exercising a variety of control strategies. The system also provides tools for simulating test data and for comparing the relative average performance under different strategies.

Keywords: control strategies, probabilistic evidential reasoning, inference network, expert systems.

## Acknowledgment
This study was partially supported by the National Science Foundation grant #NSF DCR83-13875

## 1. Problem and Knowledge Representaton

Problem and knowledge representation for evidential reasoning tasks may be based on uncertain hierarchical inference networks. An example of a network from the MEDAS expert system [1] is shown in Figure 1. In such networks leaf nodes typically represent observable events (indicators), while higher level nodes represent events (hypotheses) whose truth may be inferred from other nodes in the tree; typically in lower levels but not necessarily. Node values represent our belief in the validity of the corresponding event. To simplify the discussion we will assume that all nodes represent binary events (true = 1, false = 0); and hence one value will be sufficient to describe our degree of belief in the event represented by node $E_i$. In probabilistic terms, $P(E_i)$ will denote the probability that the event is true. A link between nodes $E_i$ and $E_j$ represents evidential relevancy between the two corresponding events. Each link is assigned value(s) that represent the degree of significance for inferring $E_i$ from $E_j$. Once an observable indicator is reported we propagate the evidence that it carries along the network links to determine its impact on our belief in the validity of the hypotheses. Methods for evidence propagation in tree-like probabilistic inference networks, first appeared in the context of traditional decision analysis; e.g. [2] and more recently in the context of AI systems; e.g. [3] [4].

Root nodes represent the target hypotheses whose resolution is the ultimate objective of the system. Intermediate nodes may also be on the list of target hypotheses; and, in any case, we use them to form defensible argumentation of the resolution of higher level hypotheses.

Several comments, however, are in order:

1. An intermediate or top level hypothesis may sometimes be directly observable, but at a higher cost than inferring it from the observable lower level indicators.
2. We may sometimes wish to bypass low level nodes and report a value directly into an intermediate or top level node. This value is not an observation but rather a deducion which the problem solver (PS) prefers not to delineate by lower level nodes.
3. An observalbe indicator may sometimes be observed with noise, in which case its value is no longer 0 or 1, but rather in between, much the same as an intermediate node.
4. Although a hierarchical structure indicates that evidence is propagated bottom up; top-down and sideways propagation may sometimes be found very useful.

Using this framework, we may represent evidential reasoning tasks by a state space representation as follows. The state of the system at any given

144

stage is characterized by the current values on the network nodes. For the
initial state $S_0$ we assign to all top level hypotheses their prior
probabilities. The values on intermediate nodes and bottom layer nodes may
be derived from the values of their parents and the values on the links.
Observable nodes whose values were inferred rather than observed are
additionally assigned the value UNOBSERVED designated by "?". These nodes
are candidates for direct observation to be suggested for the information
acquisition process.

From the goal state point of view; the network nodes are divided into
target and non-target nodes. Target nodes represent hypotheses that need to
be resolved by the end of the process. That is, only their values take part
in the termination criteria. Goal states may be defined in a variety of ways
such as follows

$S_G^1$: The values of the top level hypotheses(is) are above or below certain
thresholds

$S_G^2$: The values of the top level hypotheses(is) and a selected group of
intermediate hypotheses are above or below certain thresholds.

$S_G^1$ would fit problems where at the final stage the only important
decision is about the top-level hypotheses and decisions about other nodes do
not lead to any operational consequences. $S_G^2$ would fit problems where the
state of intermediate nodes also impact the action plan (in addition to their
role as a mediator for higher level deductions). For instance, in medical
diagnosis of critical care disorders a node representing the state of SHOCK
(Figure 1) is an intermediate node. Yet, to device a treatment plan it is
very important to know whether the patient is or is not in SHOCK. Evidential
reasoning is the problem of transferring the network from its initial state
$G_0$ to a goal state $S_G$.

In this paper we are mainly interested in control strategies for
tree-like structures and more specifically in the information acquisition
aspects of these strategies. Uncertainty is expressed in terms of
probabilities.

2. Control Strategies

A control strategy is responsible for the following functions:

1) Termination Criterion: to decide at each stage, whether or not to
   continue gathering information in order to update the
   probability(ies) of the target node(s).

145

2) Decision Function: if a termination decision has been made, a
   decision about the value of the target node(s) must be made or
   recommended - i.e. to map the posterior probabilities of these nodes
   into a final set of decisions such as {+,-,?}.
3) Information Acquisition Policy: if the termination criterion has not
   yet been met, the strategy should supply rules for comparing the
   observable nodes in order to determine the one to be queried
   next.

Control strategies for traditional Bayesian models were extensiely investigated in statistical pattern recognition theory e.g. [5] [6]. Control strategies for hierarchical problems were developed for MEDAS and PROSPECTOR. In this paper we propose a general framework for investigating such strategies for the case of a tree-like hierarchical inference network.

In order to bypass the discussions abut utility functions, and to minimize the number of parameters to be compared, let us initially assume that at the final stage, the only meaningful decision is about the root node $N_1$, and that decisions abut other intermediate nodes do not lead to any operational consequences. That is the goal state is determined by the value of $N_1$ only and intermediate nodes only serve as a conceptual tool for a clearer presentation of the inference network and its states, and to enable generating explanations.

Control strategies for hierarchical inference networks differ from each other in the degree to which they consider the values of the nodes in various levels. On one end of the spectrum of control strategies lies the strategy that completely ignores the intermediate nodes. This strategy does not 'see' the original tree, but rather its transformation into a 2-level tree. Such a transformation is achieved by calculating the conditional probabilities of a leaf node given the root node i.e. chaining the conditional probabilities of the links that lead from that leaf node to the root [2]. This process forms direct virtual links between the leaf nodes and the root node (Figure 2). This approach presents two problems:

1. Since the net structure is taken into consideration only implicitly; the order of queries suggested by such a strategy might look peculiar or arbitrary to a person who tries to see the direction in which the inference process is proceeding based on the net structure. (This deficiency is, of course, more emphasized when we attribute meaning to the decisions about the intermediate nodes).

146

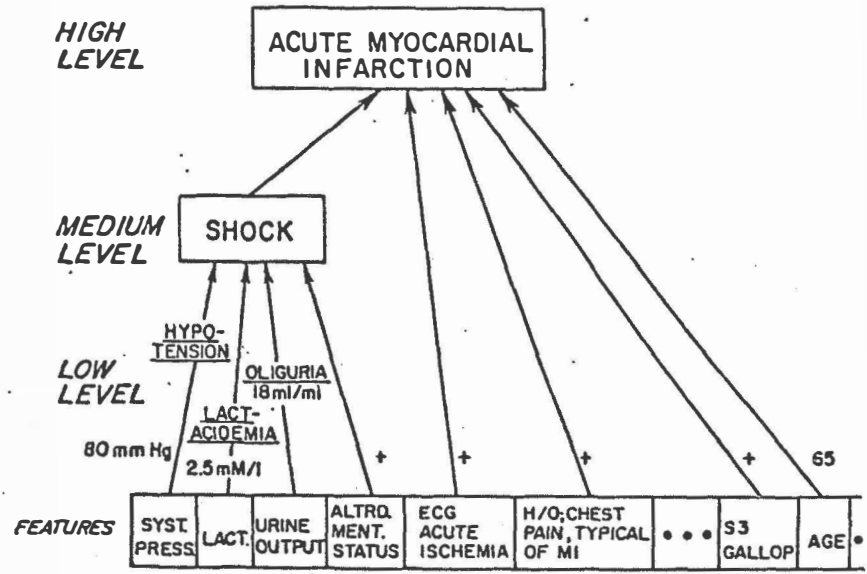

Figure 1: HIERARCHIAL STRUCTURE OF DISORDERS

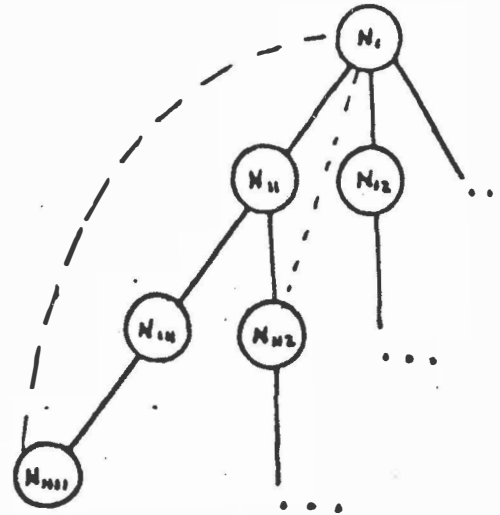

Figure 2: Virtual links are indicated by broken lines.

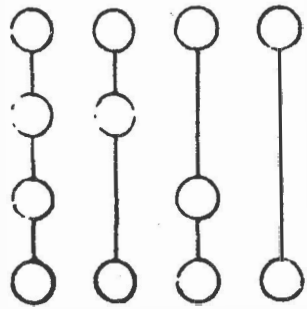

Figure 3: Four ways to look at a 4-level linkage

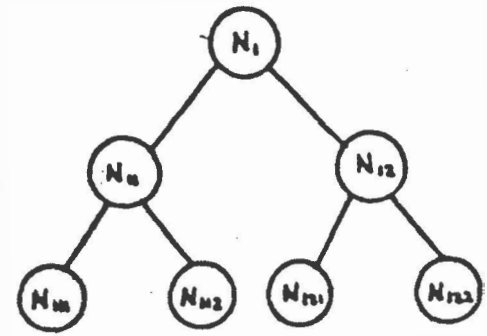

Figure 4: A 3-level tree).

2. The transformation into a 2-level tree places quite a burden on the computational resources because it has to be recalculated after every step.

The severity of this problem increases as the tree gets to be wider and deeper. We can overcome these problems by utilizing the net structure in two ways that complement each other:

1. By a staged look-ahead policy that compresses several layers into one layer. This is done by creating virtual links between a node at level k and its ancestor nodes at level $k + k'$.
2. By limiting the scope for any give stage. That is; by focusing on a small subtree and setting its root as a temporary subgoal for the next immediate stages.

Staged look-ahead is achieved by replacing a jump that goes all the way up by smaller jumps that skip only a few levels at a time. In a 3-level tree there exists only one level that we may possibly ignore, therefore there are two ways to refer to the tree (2 or 3 levels). A 4-level linkage can be transformed into two different 3-level links and into one 2-level link, as in Figure 3.

These four variations of the linkage might be represented by the following four Depth Vectors:

    a : [1,1,1]
    b : [1,2]
    c : [2,1]
    d : [3] .

The first element of each vector represents the depth of 'looking down' the tree from the root node, while ignoring intermediate levels. The next element represents the depth of 'looking down' from the level reached by the previous element, etc. This can alternatively be represented by a Level Vector, that represents the numbers of the referred levels (in addition to level 1):

    a' : [2,3,4]
    b' : [2,4]
    c' : [3,4]
    d' : [4] .

148

If we do not totally ignore the intermediate nodes, but rather refer to the original tree or to its transformation into a tree with fewer (but more than two) levels, we can extend the definitions of the components of the control strategies to include reference to the intermediate nodes i.e. termination critertion, decison function and information acquisition that are oriented to an intermediate node. Let us demonstrate the various degrees of consideration to intermediate nodes using the tree in Figure 4.

Without loss of generality let us assume, that at each level the nodes are ordered according to their a priori inferential influence on their parent, i.e., if the function $EV(N_i, N_j)$ represents the potential influence of $N_j$ on $N_i$, then $EV(N_1, N_{11}) \geq EV(N_1, N_{12})$, $EV(N_{11}, N_{111}) > EV(N_{11}, N_{112})$, etc. We also assume that EV is a fixed evaluation function, i.e. one that does not depend on the prior probability of the parent node.

If we ignore the intermediate level, the indicators will be selected according to their effect on $N_1$. The first selected indicator might be $N_{111}$ or $N_{121}$, depending on the parameters of the virtual link between these nodes and $N_1$. Ignoring the intermediate level will only preserve the order among indicators that belong to the same parent node. A possible sequence of selected indicators might be: $N_{111}, N_{121}, N_{122}, N_{112}, N_{113}, N_{123}$.

For human-engineering considerations, we might want to check the indicators in groups, i.e. initially focus on an intermediate node, then select its indicators, and only after we are through with that node move to another intermediate node and its indicators, etc. In such a manner we substitute the global information acquisition criterion (influence of the indicators on $N_1$) by a criterion that is composed of two local criteria: the influence of the intermediate node on $N_1$, and that of the observable node (indicator) on the intermediate node. A typical sequence of selected indicators might be: $N_{111}, N_{112}, N_{113}, N_{121}, \ldots$ . In such a strategy, as well as in the previous one, the termination criterion considers only $P(N_1)$, thus it is not necessary to attribute probabilities to the intermediate nodes, but merely to recognize the probabilistic links with the root and with their children.

An additional degree of consideration is achieved by distributing the termination criterion. In order to do so, after each iteration we must update, not only the probability of $N_1$, but also that of the intermediate node. If the termination criterion of, say $N_{11}$, has been met, we abandon $N_{11}$ and turn to $N_{12}$. If in the meanwhile the criterion for $N_1$ has been met, the whole process is terminated. Such a strategy needs to supply parameters to
each node in order to define its own termination criterion.

A higher degree of intermediate level consideration is achieved by
complete isolation of the termination criteria on the various levels. In
such a way the intermediate node is temporarily set as the new sub goal, and
the original goal is ignored until the solution of the new one is terminated.
This isolation may often result in wasteful behavior; the process is
continued until some intermediate level is solved, even though the
accumulated information is sufficient for solving $N_1$. The advantage of this
strategy is that we do not have to update $P(N_1)$ after each observation,
but only after terminating the solution of the intermediate node. This may
not be relevant in our model, but in extended models of distributed inference
the difference might be significant if prices are set for inter-node
communications.

3. <u>INFERNET1 - A System to Test the Strategies</u>

To test the behavior of the various strategies a computer system
INFERNET1 was implemented that offers the following functions:

1. Definition of various inference networks, and of elementary
   relational queries on these networks.
2. Management of interactive evidential reasoning e.g. diagnosis
   sessions based on a given network.
3. Generation of random data files according to a given network.
4. Diagnosis of patterns from the above mentioned data files
   using various control strategies.
5. Statistical analysis of the data collected during the offline
   diagnosis in order to measure the performance and cost of various
   strategies.

INFERNET1 was implemented in C-Prolog running under Unix on a DEC VAX
computer. In a forthcoming article we will report the findings of extensive
experiments with INFERENET1.

5. <u>Summary</u>

In this article we formulated control strategies that utilize staged
look-ahead and temporary focus on subgoals using the concept of Depth Vector,
that essentially represents a condensed tree in which only the nodes that are
considered by the control strategy are included. To improve human-machine communication; we would like to implement in expert systems control strategies that utilize these means since they are more human-oriented. A key issue is how much, if at all, efficiency we sacrifice by giving up mathematically-oriented strategies. A partial answer to this issue will hopefully be given by the experiments with INFERNET1.